# Validation of artificial neural networks to model the acoustic behaviour of induction motors

F.J. Jiménez-Romero [a,*,1], D. Guijo-Rubio [b,1], F.R. Lara-Raya [a], A. Ruiz-González [c], C. Hervás-Martínez [b]

[a] *Department of Electrical Engineering, Universidad de Córdoba, Córdoba, Spain*
[b] *Department of Computer Science and Numerical Analysis, Universidad de Córdoba, Córdoba, Spain*
[c] *Department of Electrical Engineering, University of Málaga EII, Málaga, Spain*

**ABSTRACT**

In the last decade, the sound quality of electric induction motors is a hot topic in the research field. Specially, due to its high number of applications, the population is exposed to physical and psychological discomfort caused by the noise emission. Therefore, it is necessary to minimise its psychological impact on the population. In this way, the main goal of this work is to evaluate the use of multitask artificial neural networks as a modelling technique for simultaneously predicting psychoacoustic parameters of induction motors. Several inputs are used, such as, the electrical magnitudes of the motor power signal and the number of poles, instead of separating the noise of the electric motor from the environmental noise. Two different kind of artificial neural networks are proposed to evaluate the acoustic quality of induction motors, by using the equivalent sound pressure, the loudness, the roughness and the sharpness as outputs. Concretely, two different topologies have been considered: simple models and more complex models. The former are more interpretable, while the later lead to higher accuracy at the cost of hiding the cause-effect relationship. Focusing on the simple interpretable models, product unit neural networks achieved the best results: 38.77 for MSE and 13.11 for SEP. The main benefit of this product unit model is its simplicity, since only 10 inputs variables are used, outlining the effective transfer mechanism of multitask artificial neural networks to extract common features of multiple tasks. Finally, a deep analysis of the acoustic quality of induction motors in done using the best product unit neural networks.

© 2020 Elsevier Ltd. All rights reserved.

## 1. Introduction

Electric induction motors are used in a wide range of industrial and household applications, from small electrical devices, to large industrial machinery and transport vehicles.

When an induction motor is designed, it is optimized to work powered by a 50 Hz sinusoidal signal. In these conditions, the motor generates the lowest level of electromagnetic noise. Therefore, the noise increases if the motor is fed by a non-sinusoidal signal, for instance, when a power inverter is used to generate the feed signal, using Pulse Width Modulation (PWM) techniques. This sort of technique is widely used to control the operation of the induction machine, emitting a higher noise [1,2].

Three noise components can be distinguished according to their source: mechanical, aerodynamic and electromagnetic noise. Specifically, the mechanical noise is the result of friction in the shaft bearings, whereas, aerodynamic noise is caused by the flow of air driven by the fan through the machine. On the other hand, the electromagnetic component is originated by the interactions of the electromagnetic fields generated in the stator and rotor.

Due to the large number of applications in which the electric induction motor is used, the population is exposed to noise during a long time, resulting in physical and psychological discomfort. Physical discomfort caused by noise is typically measured by the equivalent sound pressure level (*Laeq*), measured in decibels (dB). The *Laeq* is the mean energy of the noise level averaged over the measurement time interval, and is considered constant over the measurement period.

The psychological impact of acoustic annoyance is studied using different psychoacoustic parameters to define noise. These parameters are Loudness (*L*), Roughness (*R*) and Sharpness (*SA*), among others.

*Laeq* is determined by the effective level measured by a microphone during the time interval of the measurement, and is converted from pascals to decibels, whereas the psychoacoustic parameters (*L*, *R* and *SA*) are determined by applying algorithms and filters to the sound signal.

Hence, it is necessary to capture the noise of the source of interest isolated from the environment, in this work, the noise emitted by an electric induction motor, so as not to interfere with other sources of noise, through the use of appropriate acoustic equipment.

An alternative approach to studying noise is through the development of empirical models, using Artificial Neural Networks (ANNs) [3]. One of the main benefits of these ANN models is their ability to learn multiple related tasks [4,5]. This sort of learning is also known as multitask learning, which allow predicting several outputs simultaneously in a single model, with many fewer connections between nodes and low complexity compared to single models for every output. Finally, as a novelty, product unit and sigmoid units ANNs are used, analysing their performance on the prediction of the acoustic quality of the induction motor. Techniques based on artificial neural networks are widely used in process engineering, due to their high precision to model complex non-linear systems with multiple input and output variables [6,7].

In acoustics, due to the non-linear nature of sound, these techniques are often used for various purposes, such as noise prediction, analysis and evaluation. Some of these works use ANNs to classify noise inside and outside vehicles by recording the noise and then compare the results of the proposed models with the subjective results of a jury [8,9]. Other authors also investigate the relationship between subjective and objective classification, using psychoacoustic parameters such as *L* [6]. Concretely, the subjective section of the study is carried out by a jury, whereas, the objective section is done by using ANNs.

Some works focus exclusively on the objective assessment of sound quality using ANN models, studying the noise emitted by the design of portable measuring devices for general use [10]. Moreover, [11] studied the use of ANNs with the purpose of predicting the acoustic quality emitted by a combustion engine with promising results. Also, Wang et al. [12] analysed the sound and thermal isolation properties of ultra-fine glass wool mats at different frequencies using ANN. Furthermore, ANNs have been used to develop a new technique to produce fast and robust auralizations for room acoustics simulation [13].

The noise generated by electric motors powered by inverters has been modelled in two ways: 1) In [14], by the electric power signal and the mechanical vibration, whereas, 2) in [15] it was done by recording the noise of an electric motor isolated in a semi-anechoic chamber.

Based on the limitations of previous works, it would be interesting to obtain an objective value of the psychoacoustic parameters (*L*, *R* and *SA*) and the *Laeq* of an electric induction motor, using a selection of constructive and electrical signal feed parameters involved in the generation of the noise emitted by the machine. The principal advantage of using these parameters is that, it is not necessary to separate the noise of the electric motor from the environmental noise. Besides, it is much easier to measure the electrical parameters, than to register the sound emitted by the motor, and then, calculate the sound quality parameters.

The main goal of this work is to use ANNs to obtain an empirical multivariate model, interpretable and not computational intensive, i.e. simple models able to achieve a competitive performance considering a small number of input variables, in order to predict the acoustic behaviour of induction motors successfully without the need of a huge amount of information. The model simultaneously estimates the *Laeq*, *L*, *R* and *SA* of an electric induction motor fed by a power inverter. The input variables used are: the motor speed, the modulation index of the PWM technique, the total harmonic distortion of the voltage and current waveforms, and the frequency spectra of the voltage and current signals. Alternatively, more complex ANNs have also been analysed, increasing both, the number of nodes in the hidden layer and the number of inputs considered by the models. Concretely, these models are more accurate and improve the results achieved by the simple models. Therefore, a decision should be made: 1) simple, interpretable and computationally efficient models, and 2) more complex and computationally intensive ANNs, being able to achieve higher accuracy.

The remainder of this paper is organized as follows: In Section 2, the materials used to carry out the experiments as well as the dataset considered are introduced and explained, detailing the process for final data acquisition. Moreover, in this section, the ANN methodology is described. The results achieved are shown in Section 3. Furthermore, an in-depth analysis of the best model obtained and the acoustic quality of the induction motor are presented in Section 4. Finally, Section 5 concludes the paper with some final remarks.

## 2. Material and methods

### 2.1. Experimental setup

The motor used in our experimental setup is a three-phase induction motor with a dahlander-type squirrel cage, which can be used to select the number of poles, i. e. it allows to change the turning speed without modifying the constructive parameters.

The tests were performed inside a semi-anechoic chamber, complying with the ISO 3745 standard, for frequencies between 100 Hz and 10 kHz, to avoid interferences in the measurements of sound.

The acoustic measurement equipment used in our study consisted of the following components:

- *Sinus Soundbook* multichannel analyser integrated in a *PC Panasonic CF-18 Toughbook*; the analyser is an acoustic measurement and analysis device with up to 8 high resolution 20-bit measurement channels.
- Software *Samurai Version 1.7.14 © 2005 Sinus Messtechnik GmbH*, which is used in *Sinus Soundbook* devices and *Harmonie* analyser.
- Pre-amplified microphone set, model *G.R.A.S. 46AE*, developed by *Sound & Vibration A/S*; this set consists of a microphone, model *G.R.A.S. 40AE*, and *G.R.A.S. 26CA* pre-amplifier.
- Acoustic calibrator, model 4231 developed by *Brüel & Kjaer*; the calibration frequency of the calibrator is 1000 Hz with a sound pressure of 94 ± 0.2 dB. In addition, it includes a second calibration signal with the same frequency, but a sound pressure of 120 ± 0.2 dB.

Furthermore, the power equipment is a *Semikron Semitech* inverter model *SKM 50 GB 123D*, with maximum working values of 400 V and 30 A, and a switching frequency of 20 kHz; it consists of a three-phase full-wave rectifier bridge, model *B6U* developed by *Semikron*, and an insulated-gate bipolar transistor three-phase bridge, model *B6CI* developed by *Semikron* as well.

The instrument used to record the electrical measurements is a *435-Series II* developed by *Fluke*, which has 4 voltage and 4 power inputs with a sampling rate of 256 samples per cycle per input. Harmonic and inter-harmonic groups from 1 to 50 based on the standard *IEC 61000-4-7* are considered.

The control equipment used in our study is responsible for controlling the inverter, loading the control vectors from the application, and applying the signals of the different techniques to be tested. This equipment consists of a microcontroller mounted on prototyping *PCB*, optocoupler circuit with *4n35* integrated circuits, and software for the generation and control of the PWM techniques created using *Labview 2009* with the National Instrument graphic programming language.

As previously mentioned, all tests were conducted inside a semi-anechoic chamber, with microphones attached to tripods directed at the centre of the envelope of the motor. The induction motor was supplied through the modulated inverter with the different PWM techniques designed for different objectives [16–18] and 50 Hz for all tests. To prevent any magnetic saturation of the motor, all measurements were recorded for voltage values under 80% of the nominal value of the machine. Fig. 1 shows the measurement setup used for data collection in our tests.

The microphones are calibrated using the acoustic calibrator 4231, developed by *Brüel* & *Kjaer*. In particular, the microphones were inserted one by one into the calibrator and were automatically calibrated using the *Samurai* software, installed in the acoustic measurement equipment, the *Sinus Soundbook*. Before each test, the microphones are calibrated, and the distances of the microphones and the motor inside the semi-anechoic chamber are verified.

Once the microphones are calibrated, the different software applications are executed in this order: 1) the *Labview* application for the generation of the control signal. 2) *Flukeview* software for the collection of data from the measurement equipment and transferring them to the PC. And, 3) *Samurai* software for the collection of the sound recorded by the microphones. The next step is to connect the adjustable power supply and ensure that the output voltage is correct, being necessary for the supply and control of the drivers of the inverter.

Moreover, the number of poles connected to the motor is checked, to ensure its suitability for the test being conducted. Then, the inverter is supplied, the test to be implemented is selected in the *Labview* application (PWM technique and parameter values), and finally, the test is executed. In this way, the motor starts, and then, when the intensity harmonics is stabilized (checked using the *Fluke 435* measurement equipment), the microphone measurements are recorded for 10 s, using the *Soundbook* measurement device.

### 2.2. Inputs and output description

The aerodynamic component is the most influential on the level of noise generated, depending on the rotation speed of the engine. The noise emission is caused by the airflow passage of the cooling system and the siren effect of the fan blades.

The speed of the motor synchronism depends on both, the frequency of the supply signal, and the number of poles, $p$. For this work, a fixed supply signal frequency of 50 Hz and a number of poles $p = \{2, 4, 6, 12\}$ are used as input. The synchronism speed of the induction motor, $Ns$, is determined by the expression $Ns = \frac{60f}{2p}$, thus, the higher the number of poles, the lower the synchronism speed. Therefore, following the previous expression, the $Ns$ values corresponding to those values of $p$ are $Ns = \{3000, 1500, 1000, 500\}$. The number of poles is a constructive parameter of the induction motor, and in order to use different numbers of poles, a four-speed dhalander motor has been used.

The mechanical component of the noise emitted is the result of friction in bearings and misalignment of the rotor shaft; therefore it depends on the speed of the motor. The mechanical component is the least contributor to the overall noise level.

The electrical parameters of the power supply signal directly influence on the electromagnetic component due to its electrical nature. The origin of the noise is due to the interaction of the electromagnetic inductions, created by the supply signal. The electrical parameters of the supply signal are defined by the Modulation index ($M$) of the PWM techniques, the harmonic distortions of voltage ($Vthd$) and intensity ($Ithd$), and by the power level of the fundamental harmonics and subharmonics shown in Table 1. For the selection of the most representative voltage subharmonics, those with a value higher than 5% of the fundamental harmonic of 50 Hz of voltage have been identified. Whereas, in the case of current harmonics, those with a value higher than 1.5% of the fundamental harmonic of 50% of intensity have been identified. To improve the readability, Table 2 shows the nomenclature used for the diverse input variables.

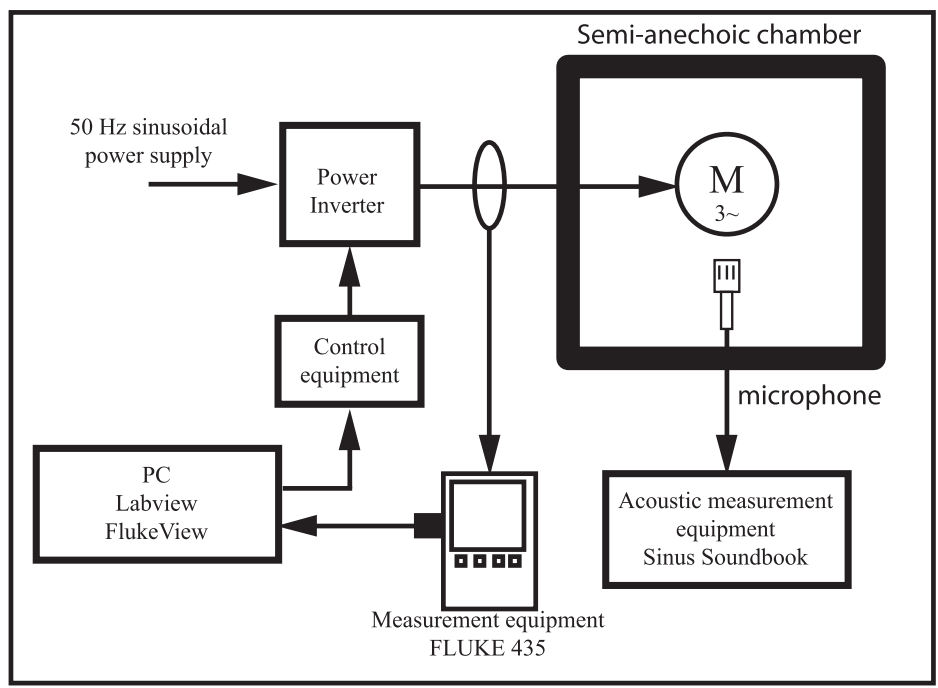

**Fig. 1.** Experimental setup.

In the field of psychoacoustics, acoustic quality parameters are used to quantify how soothing or disturbing a given sound is. In this sense, four different parameters are considered: 1) $Laeq$ determines the intensity of sound generated by sound pressure, 2) $L$ is a standardized subjective measure of the intensity with which the human ear perceives a sound, 3) $R$ quantifies the level of disturbance perceived by the ear due to quick fluctuations, which has the highest level when a sound is modulated at 70 Hz, and, finally, 4) $SA$ is used to measure the high-frequency content of a sound (because this parameter is not standardized, it can be calculated using different methods [19]).

### 2.3. Experimental data

To regulate the supply of the electrical machines, several modulation techniques were employed in this study. These include Harmonics Injection Pulse Width Modulation Frequency Modulated Triangular Carrier (HIPWM-FMTC) [17], which is a pulse width modulation technique with harmonic injection into the modulating waveform, and frequency modulated by a sinusoidal function –or linear, in the case of HIPWM-FMTC2 [18]– of the carrier waveform. In this technique, $k_c$ and $f_c$ parameters allow the modulation of the triangular carrier to be modified. In contrast, Slope Pulse Width Modulation (SLPWM) [16] uses a trapezoidal wave to define the modulating wave, as well as, a discontinuous sinusoidal wave to generate the carrier wave. The carrier frequency is controlled using the parameter $k$, (in particular, $k\mu$ times of the modulating waveform, where $\mu$ is another control parameter), that allows the adjusting of the frequency modulation order.

All these techniques have diverse objectives, for example, the primary objective of the HIPWM-FMTC technique is to achieve good wave quality concerning harmonic distortion. Otherwise, the SLPWM technique also aims to achieve good wave quality, albeit using lower modulation indexes. Furthermore, the HIPWM-FMTC2 technique is aimed to reduce the noise originated inside an electric motor. To obtain a representative sample of tests and consider a wide variety of values for each control parameter, diverse values for $M$ and $p$ were considered, specifically $M = \{5, 7, 9, 11, 13, 15, 17, 19, 21\}$ and $p = \{2, 4, 6, 12\}$.

The values selected for the control parameters of the SLPWM technique are listed in Table 3, and each of these combinations was used for the different values of $p$, resulting in a total of 396 patterns.

On the other hand, the control parameters of the HIPWM-FMTC technique include $k_c$ and $f_c$. The combinations of these control parameters in the modulation technique have been selected to obtain the diverse values of $M$. Table 4 shows the maximum and minimum values for $k_c$ and $f_c$, as well as, their increases and the number of tests for a given value of $M$. With these combinations and the different values of $p$, 2272 patterns were obtained.

In contrast, for the HIPWM-FMTC2 technique, the control parameter is the angle $\alpha$, which varies between 17° and 45°, with increments of 1°. The number of tests per value of $M$ is 29 with 9 and 4 possible combinations based on $M$ and $p$, respectively. Therefore, 1044 patterns were performed in total.

To sum up, 396 patterns were obtained from SLPWM, 2272 from HIPWM-FMTC and other 1044 from HIPWM-FMTC2. Thus, the dataset considered in this paper is made up of 3712. Then, this dataset is divided into two different sets: 2784 (75%) patterns for the training set, whereas the remaining 928 (25%) patterns belongs to the test set, following the guidelines given by [20]. The working ranges of the input and output variables are shown in the Table 5.

**Table 1**
Percentage value of the voltage and current harmonics compared to the fundamental 50 Hz harmonic.

| Hz | (%)$V50$ | (%)$I50$ | Hz | (%)$V50$ | (%)$I50$ |
|---|---|---|---|---|---|
| 50 | 100.00 | 100.00 | 1250 | 11.05 | 5.90 |
| 100 | < 5% | 2.87 | 1450 | 10.14 | 4.70 |
| 200 | < 5% | 2.25 | 1550 | 8.59 | 3.68 |
| 250 | 8.19 | 19.35 | 1750 | 8.02 | 3.05 |
| 350 | 8.05 | 14.19 | 1850 | 8.36 | 2.93 |
| 550 | 8.72 | 10.55 | 2050 | 7.73 | 2.47 |
| 650 | 11.43 | 11.48 | 2150 | 7.40 | 2.17 |
| 850 | 11.50 | 9.04 | 2350 | 6.77 | 1.86 |
| 950 | 9.83 | 6.88 | 2450 | 7.00 | 1.74 |
| 1150 | 9.34 | 5.49 | | | |

**Table 2**
Nomenclature of the input variables.

| | | | | | | | |
|---|---|---|---|---|---|---|---|
| $X_1$ | $Vthd$ | $X_{11}$ | $V950$ | $X_{21}$ | $V2450$ | $X_{31}$ | $I1150$ |
| $X_2$ | $Ithd$ | $X_{12}$ | $V1150$ | $X_{22}$ | $I50$ | $X_{32}$ | $I1250$ |
| $X_3$ | $p$ | $X_{13}$ | $V1250$ | $X_{23}$ | $I100$ | $X_{33}$ | $I1450$ |
| $X_4$ | $M$ | $X_{14}$ | $V1450$ | $X_{24}$ | $I200$ | $X_{34}$ | $I1550$ |
| $X_5$ | $V50$ | $X_{15}$ | $V1550$ | $X_{25}$ | $I250$ | $X_{35}$ | $I1750$ |
| $X_6$ | $V250$ | $X_{16}$ | $V1750$ | $X_{26}$ | $I350$ | $X_{36}$ | $I1850$ |
| $X_7$ | $V350$ | $X_{17}$ | $V1850$ | $X_{27}$ | $I550$ | $X_{37}$ | $I2050$ |
| $X_8$ | $V550$ | $X_{18}$ | $V2050$ | $X_{28}$ | $I650$ | $X_{38}$ | $I2150$ |
| $X_9$ | $V650$ | $X_{19}$ | $V2150$ | $X_{29}$ | $I850$ | $X_{39}$ | $I2350$ |
| $X_{10}$ | $V850$ | $X_{20}$ | $V2350$ | $X_{30}$ | $I950$ | $X_{40}$ | $I2450$ |

**Table 3**
Values assigned to the control parameter $k$ of the SLPWM technique.

| $M$ | $k$ (SLPWM) |
|---|---|
| 5 | −0.74, −0.84, −0.94, −1.04, −1.14, −1.24, −1.34, −1.44, −1.54, −1.64, −1.74 |
| 7 | 1.24, 1.34, 1.44, 1.54, 1.64, 1.74, 1.84, 1.94, 2.04, 2.14, 2.24 |
| 9 | −2.74, −2.64, −2.54, −2.44, −2.34, −2.24, −2.14, −2.04, −1.94, −1.84, −1.75 |
| 11 | 2.25, 2.35, 2.45, 2.55, 2.65, 2.75, 2.85, 2.95, 3.05, 3.15, 3.24 |
| 13 | −3.74, −3.64, −3.54, −3.44, −3.34, −3.24, −3.14, −3.04, −2.94, −2.84, −2.75 |
| 15 | 3.25, 3.35, 3.45, 3.55, 3.65, 3.75, 3.85, 3.95, 4.05, 4.15, 4.24 |
| 17 | −4.74, −4.64, −4.54, −4.44, −4.34, −4.24, −4.14, −4.04, −3.94, −3.84, −3.75 |
| 19 | 4.25, 4.35, 4.45, 4.55, 4.65, 4.75, 4.85, 4.95, 5.05, 5.15, 5.24 |
| 21 | −5.74, −5.64, −5.54, −5.44, −5.34, −5.24, −5.14, −5.04, −4.94, −4.84, −4.75 |

**Table 4**
Values assigned to the control parameters $k_c$ and $f_c$ for the HIPWM-FMTC modulation technique.

| M | $k_c$ (min – max) | $f_c$ (min – max) | $\triangle k_c$ | $\triangle f_c$ | Tests |
|---|---|---|---|---|---|
| 5 | 0–10 | 5–10 | 0.25 | 0.125 | 41 |
| 7 | 0–14 | 7–14 | 0.25 | 0.125 | 57 |
| 9 | 0–18 | 9–18 | 0.30 | 0.15 | 61 |
| 11 | 0–22 | 11–22 | 0.35 | 0.175 | 64 |
| 13 | 0–26 | 13–26 | 0.50 | 0.25 | 53 |
| 15 | 0–30 | 15–30 | 0.50 | 0.25 | 61 |
| 17 | 0–34 | 17–34 | 0.50 | 0.25 | 69 |
| 19 | 0–38 | 19–38 | 0.50 | 0.25 | 77 |
| 21 | 0–42 | 21–42 | 0.50 | 0.25 | 85 |

### 2.4. Artificial neural network models (ANNs)

ANNs [3] are a widely used kind of non-linear models due to its potential to solve a vast range of problems. Feed-forward Neural Networks were one of the first proposals made in the literature; they were composed by at least one input layer, one hidden layer and one output layer, using in the hidden layer different sorts of basis functions [3], with diverse goals depending on the problem tackled. Furthermore, ANNs have been proven to be universal approximators with the difference in the basis function used for each hidden layer neuron.

If a training dataset $D = (\mathbf{x}_i, y_i); i = \{1, 2, \ldots, n\}$ is available, where $\mathbf{x}_i = (x_{1i}, \ldots, x_{di})$ is the vector of measurements taking values in $\mathbf{x}_i \in \mathbb{R}^d$, and $y_i \in \mathbb{R}^r$; then regardless of the basis functions of the hidden layer, ANN can be written as:

$$f_k(\mathbf{x}, \mathbf{w}, \boldsymbol{\beta}) = \sum_{j=1}^{m} \beta_{kj} B_j(\mathbf{x}, \mathbf{w}_j), \qquad k = 1, \ldots, r \qquad (1)$$

where $B_j(\mathbf{x}, \mathbf{w}_j)$ represents a set of non-linear transformations of the input vector $\mathbf{x}^T = (x_1, x_2, \ldots, x_d)$, with $d$ being its length; $r$ is the number of outputs; bias is considered with the element $B_0(\mathbf{x}, \mathbf{w}_j) = 1$; $\boldsymbol{\beta}_k^T = (\beta_{k1}, \beta_{k2}, \ldots, \beta_{km})$ are the coefficients from linear combination estimated from the data; $\mathbf{w}_j^T = (w_{j1}, w_{j2}, \ldots, w_{jd})$ are the parameters related to the basis function; $m$ is the number of basis functions required to minimise some definite function of the error.

In the problem considered, the input layer is composed of $d = 40$ neurons, one for each input variable, whereas the output layer is composed of $r = 4$ neurons, associated to the different variables previously commented in Sections 2.2 and 2.3.

Regarding to the hidden layer, we have focused on the following two basis functions being the most commonly applied in the state-of-the-art:

- Sigmoid unit [21] is the most used basis function due to its ability to approximate any continuous function accurately. However, they fail in local optimum frequently. Using the notation described above, the sigmoid unit is represented as:

$$B_j(\mathbf{x}, \mathbf{w}_j) = \frac{1}{1 + e^{-\left(w_{j0} + \sum_{i=1}^{d} w_{ji} x_i\right)}}, \qquad j = 1, \ldots, m \qquad (2)$$

- Product unit [22] is a basis function that not only is it able to retain the properties of a universal approximator, but also, it only uses a small number of multiplicative neurons [23]. According to the notation followed so far, the product unit is expressed as:

$$B_j(\mathbf{x}, \mathbf{w}_j) = \prod_{i=1}^{d} x_i^{w_{ji}}, \qquad j = 1, \ldots, m \qquad (3)$$

Owing to their architecture, Product Unit Neural Network (PUNN) models provide high versatility for implementing high-order functions, retaining the properties of a universal approximator and using only a small number of neurons with multiplicative units instead of addictive ones [23]. On the other hand, Sigmoid Unit Neural Network (SUNN) models use sigmoid transfer functions for nodes of the hidden layer. This kind of neural network is most widely used owing to its ability to approximate any continuous function accurately. As PUNN and SUNN do not require a high number of neurons for solving certain problems [24], their application to this problem is reasonable.

Initially, these kinds of models were trained using backpropagation algorithms [25,26], involving a notable disadvantage: they usually do not determine the global minimum of the error function, but only a local minimum. Thus, a metaheuristic such as an evolutionary algorithm [27] is required. Further details of this algorithm could be found in [28]. The parameters of the evolutionary algorithm are shown in Table 6, whereas, the remaining values were obtained from [29].

### 2.5. ANN evaluation

The ANN models applied to regression problems are frequently evaluated in terms of Mean Squared Error (MSE) and the Standard Error of Prediction. In this sense, MSE is expressed by Eq. 4:

$$MSE = \frac{1}{r} \frac{1}{n} \sum_{q=1}^{r} \sum_{i=1}^{n} (y_i^q - \hat{y}_i^q)^2, \qquad (4)$$

where $r$ is the number of outputs considered, in this case 4, $n$ is the size of the dataset (2784 for the training set and 928 for the generalisation set), and $y_i^q$ and $\hat{y}_i^q$ are the real and predicted values of output $q$, respectively. Moreover, SEP is defined by Eq. 5:

**Table 5**
Working ranges of the input and output variables.

| | | | | | | | | | |
|---|---|---|---|---|---|---|---|---|---|
| $X_1$ (%) | 6.70/229.60 | $X_{11}$ (V) | 0.02/92.70 | $X_{21}$ (V) | $8.0e^{-3}$/68.30 | $X_{31}$ (A) | $4.6e^{-5}$/0.07 |
| $X_2$ (%) | 1.80/411.10 | $X_{12}$ (V) | 0.02/82.90 | $X_{22}$ (A) | 0.04/0.38 | $X_{32}$ (A) | $2.1e^{-5}$/0.07 |
| $X_3$ (p) | 2.00/12.00 | $X_{13}$ (V) | 0.06/80.30 | $X_{23}$ (A) | $4.4e^{-5}$/0.05 | $X_{33}$ (A) | $4.0e^{-5}$/0.07 |
| $X_4$ (M) | 5.00/21.00 | $X_{14}$ (V) | 0.03/86.30 | $X_{24}$ (A) | $4.4e^{-5}$/0.02 | $X_{34}$ (A) | $3.6e^{-5}$/0.05 |
| $X_5$ (V) | 0.32/244.60 | $X_{15}$ (V) | 0.03/69.40 | $X_{25}$ (A) | $7.0e^{-5}$/0.50 | $X_{35}$ (A) | $2.6e^{-5}$/0.05 |
| $X_6$ (V) | 0.02/121.40 | $X_{16}$ (V) | $2.0e^{-3}$/72.70 | $X_{26}$ (A) | $1.2e^{-4}$/0.30 | $X_{36}$ (A) | $2.7e^{-5}$/0.05 |
| $X_7$ (V) | 0.02/95.20 | $X_{17}$ (V) | $1.0e^{-3}$/78.90 | $X_{27}$ (A) | $1.6e^{-4}$/0.17 | $X_{37}$ (A) | $2.3e^{-5}$/0.04 |
| $X_8$ (V) | 0.03/87.60 | $X_{18}$ (V) | $3.0e^{-3}$/67.00 | $X_{28}$ (A) | $5.1e^{-5}$/0.17 | $X_{38}$ (A) | $2.7e^{-5}$/0.03 |
| $X_9$ (V) | 0.01/99.10 | $X_{19}$ (V) | 0.01/65.10 | $X_{29}$ (A) | $5.3e^{-5}$/0.12 | $X_{39}$ (A) | $3.7e^{-5}$/0.03 |
| $X_{10}$ (V) | 0.02/96.50 | $X_{20}$ (V) | 0.01/68.20 | $X_{30}$ (A) | $4.3e^{-5}$/0.11 | $X_{40}$ (A) | $1.7e^{-5}$/0.03 |

**Table 6**
Parameters and values of the evolutionary algorithm and PUNN and SUNN models.

| Evolutionary algorithm | |
|---|---|
| Runs | 30 |
| Generations | 200 |
| Population size | 1000 |
| Hidden nodes to be created or deleted | [1, 2] |
| Minimum hidden nodes initialisation | 1 |
| Maximum hidden nodes initialisation | 1 |
| Maximum hidden nodes whole process | 3 |
| PUNN models | |
| Weights between input-hidden layer | [−1, 1] |
| Weights between hidden-output layer | [−5, 5] |
| SUNN models | |
| Weights between input-hidden layer | [−5, 5] |
| Weights between hidden-output layer | [−5, 5] |

$$SEP = \frac{1}{r}\sum_{q=1}^{r}\frac{100}{|\bar{y}^q|}\sqrt{\frac{1}{n}\sum_{i=1}^{n}(y_i^q - \hat{y}_i^q)^2}, \quad (5)$$

where $\bar{y}^q$ is the mean value of output $q$ for the training or the generalisation set, depending on the set evaluated.

## 3. Results

In order to compare the predictive ability of ANNs, two different kind of models have been developed (PUNN and SUNN) using 40 inputs and 4 outputs: $Laeq, L, R$ and $SA$. The results are shown in Table 7.

Table 7 is divided into two parts: the first at the top shows the results obtained for the PUNN model, whereas, the second part shows the results for the SUNN model. For these two models, the mean and standard deviation (SD) of the MSE, SEP and #links, and the corresponding values for the best model are shown. To make a fair comparison, the MSE and SEP values were obtained for every output, as well as, the *Global* value, which indicates the quality of the model for both, MSE and SEP, in a general way (it is the sum of the values for every output).

Taking into account the complexity of the problem, it can be stood out that both models obtained good solutions with a small number of connections. Concretely, the best model in mean terms was the PUNN, obtaining 44.32 for MSE and 13.79 for SEP, besides, the SUNN model obtained 46.33 and 14.45, respectively. With respect to the #links, the most simple model was SUNN with 36.80 connections. Furthermore, regarding the best models, although SUNN reached to the best value in terms of MSE (38.37), the difference is slight (PUNN obtained 37.77), whereas, in terms of SEP, PUNN achieved an outstanding value of 13.11 using only 18 connections (being the most simple model found). Moreover, this best model only considered 10 variables. The architecture of this best model is shown in Fig. 2. Eqs. (6)–(9) show the output function for $\widehat{Laeq}^*, \hat{L}^*, \hat{R}^*$ and $\widehat{SA}^*$ for this best PUNN model. In general, these results outline the effective transfer mechanism of multitask ANNs to extract common features from multiple tasks.

$$\widehat{Laeq}^* = 0.192 + 1.046 \cdot B \quad (6)$$

$$\hat{L}^* = 0.318 + 0.449 \cdot B \quad (7)$$

$$\hat{R}^* = 0.234 + 0.022 \cdot B \quad (8)$$

$$\widehat{SA}^* = 0.330 + 0.131 \cdot B \quad (9)$$

where $B$ is defined as follows:

$$B = \frac{(X_4^*)^{0.016} \cdot (X_{15}^*)^{0.209} \cdot (X_{20}^*)^{0.077} \cdot (X_{26}^*)^{0.628} \cdot (X_{39}^*)^{0.088}}{(X_3^*)^{3.532} \cdot (X_5^*)^{1.415} \cdot (X_6^*)^{0.044} \cdot (X_{19}^*)^{0.016} \cdot (X_{24}^*)^{0.145}} \quad (10)$$

## 4. Discussion

In this section, an analysis of the influence of the PUNN model is provided, as well as, a detailed analysis of the acoustic quality of the induction motor using the best model obtained.

### 4.1. Analysis of the influence of the model

Using the output functions shown in Eqs. (6)–(9) for the best PUNN model, the influence of the input variables on the different outputs is analysed in Fig. 3. The higher the value of the slope, the greater the influence on the output, also, depending on whether the value is positive (coloured in blue) or negative (in red), the output will increase or decrease, respectively.

The most important variable in the generation of noise is $X_3$ (number of poles, $p$), which determines the speed of the induction motor, being responsible for the aerodynamic component of noise. Although the mechanical component is the least important in noise generation, it also depends on the number of poles ($X_3$). The variable $X_4$ (modulation index, $M$) characterises the feed signal generated by the inverter, composed of a harmonic distribution, responsible of the electromagnetic component of the noise.

**Table 7**
Results obtained for the four outputs: $Laeq, L, R$ and $SA$.

| | PUNN | | | | | | | | | | |
|---|---|---|---|---|---|---|---|---|---|---|---|
| | MSE | | | | | SEP | | | | | #Links |
| | Global | $\widehat{Laeq}$ | $\hat{L}$ | $\hat{R}$ | $\widehat{SA}$ | Global | $\widehat{Laeq}$ | $\hat{L}$ | $\hat{R}$ | $\widehat{SA}$ | |
| Mean | 44.32 | 3.06 | 41.13 | $1.83e^{-4}$ | $1.26e^{-1}$ | 13.79 | 1.51 | 4.41 | 4.76 | 3.12 | 38.80 |
| SD | 5.27 | 0.99 | 4.80 | $0.10e^{-4}$ | $1.02e^{-2}$ | 0.55 | 0.25 | 0.25 | 0.11 | 0.13 | 9.66 |
| Best | 38.77 | 1.24 | 37.39 | $1.80e^{-4}$ | $1.33e^{-1}$ | 13.11 | 0.97 | 4.21 | 4.72 | 3.21 | 18.00 |
| | SUNN | | | | | | | | | | |
| | MSE | | | | | SEP | | | | | #Links |
| | Global | $\widehat{Laeq}$ | $\hat{L}$ | $\hat{R}$ | $\widehat{SA}$ | Global | $\widehat{Laeq}$ | $\hat{L}$ | $\hat{R}$ | $\widehat{SA}$ | |
| Mean | 46.33 | 3.54 | 42.65 | $2.11e^{-4}$ | $1.35e^{-1}$ | 14.45 | 1.62 | 4.52 | 5.05 | 3.25 | 36.80 |
| SD | 7.16 | 1.00 | 6.81 | $3.77e^{-5}$ | $1.96e^{-2}$ | 0.83 | 0.23 | 0.36 | 0.44 | 0.25 | 13.92 |
| Best | 38.37 | 3.89 | 34.36 | $2.66e^{-4}$ | $1.27e^{-1}$ | 14.63 | 1.72 | 4.04 | 5.74 | 3.13 | 20.00 |

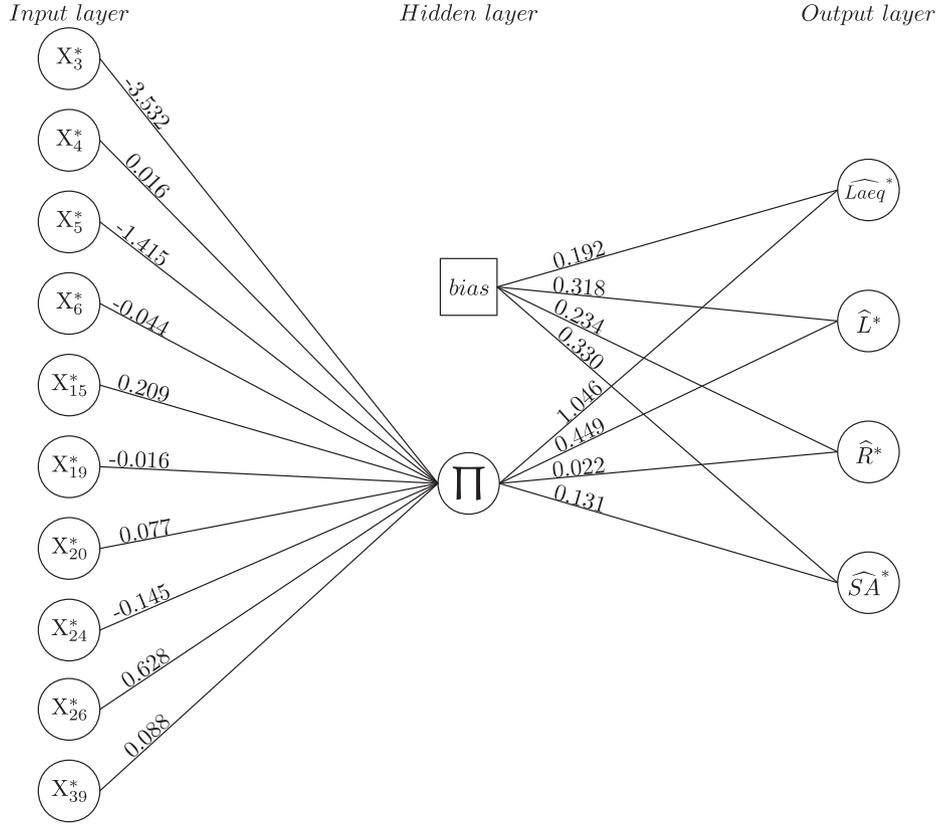

**Fig. 2.** Architecture of the PUNN model for the prediction of the acoustic quality of the induction motor.

The harmonic distribution is composed of the 50 Hz fundamental voltage harmonic ($X_5$) and a sub-harmonic group of voltage and current. From the initial selection of all the sub-harmonics (see Table 1), the algorithm selected for the PUNN model the voltage harmonics ($X_6, X_{15}, X_{19}$ and $X_{20}$, respectively, $V250, V1550, V2150$ and $V2350$) and current harmonics ($X_{24}, X_{26}$ and $X_{39}$, respectively, $I200, I350$ and $I2350$). The whole set of sub-harmonics interact with the electromagnetic fluxes generated in the rotor, stator and air gap of the induction motor.

On the other hand, the influence of the input variables on the different outputs for the SUNN model have been included in supplementary material.

### 4.2. Analysis of the acoustic quality of the induction motor

The main analysis of the acoustic quality of the induction motor is done to the most influential parameters, $p, V50, V1550$ and $I350$.

As indicated in Section 2.2, $p$ is inversely proportional to the synchronous speed of the induction motor, $Ns$. The aerodynamic component is the most influential in the generation of noise in the induction motor, therefore, the number of poles is the most important as shown in the Figs. 4a, 5a, 6a and 7a.

The electromagnetic component of the noise generated by the induction motor, is a consequence of the characteristics of the power signal. $M$ are the pulses used to construct the supply signal for the PWM technique. The lower is $M$, the higher is the harmonic distortion of the Vthd voltage and Ithd current signals, and the lower is the fundamental harmonic $V50$. Moreover, higher values of the fundamental harmonic, make the induction motor to generate less noise, as shown in Figs. 4a, 5a, 6a and 7a, because it has been designed for a 50 Hz sinusoidal feed signal. The rest of the inputs are harmonics multiples of the fundamental frequency of both, voltage and current, being a consequence of $M$.

Table 8, shows the difference between the minimum and maximum values in the graph of 35 $dB$. Studying the extreme values of both inputs, it could be observed that, for values of $p = 12$, the variation of any of the four outputs has little influence on $V50$. For values of $p$ lower than or equal to 4 poles, the values of $V50$ have influence on the different outputs.

To maintain the noise level of the induction motor at higher speeds, the magnitude of the fundamental voltage harmonic must be increased. For example, Fig. 4a, shows that for $p = 10$, a value of 25 $V$ for the 50 Hz fundamental voltage harmonic, the value of $Laeq$ is 55 dB. This value can also be maintained if $p = 8$ and the fundamental voltage harmonic increases its value to 125 $V$. With respect to the variables $V1550$ and $I250$ also have influence on the model outputs, albeit less than variables $p$ and $V50$. Figure 4b, 5b, 6b and 7b show that the variable with more influence in the outputs is $I250$ when its value increases. Although the contribution of these inputs is not very high, it would be necessary to avoid power signals that contain these harmonics to improve noise and acoustic quality.

Finally, an extensive analysis of the acoustic quality following the best SUNN model is shown in supplementary material.

### 4.3. Complex models

Even though the main objective of this paper was to obtain simple models obtaining competitive results, with the aim of demonstrating that the proposed methodology is able to improve the state-of-the-art algorithms for regression problems, a set of experiments has been carried out using the same training and test data exposed in Section 2.3.

In this way, the proposed PUNN and SUNN models are trained during a higher number of generations, concretely, 6000 generations, instead of 200 as was fixed for the initial models, whereas the remaining parameters are set to the same previous values described in Table 6. Therefore, the models obtained will be more complex in terms of both, number of connections and number of input variables considered. The results of these models are exposed in Table 9. As can be seen, SUNN obtains the best results not only in average, but also achieves the best model (29.95 and 11.46 for MSE and SEP, respectively). By contrast, according to the number of links of the best models, PUNN is the simplest with 60 links, whereas SUNN is slightly more complex (92 links).

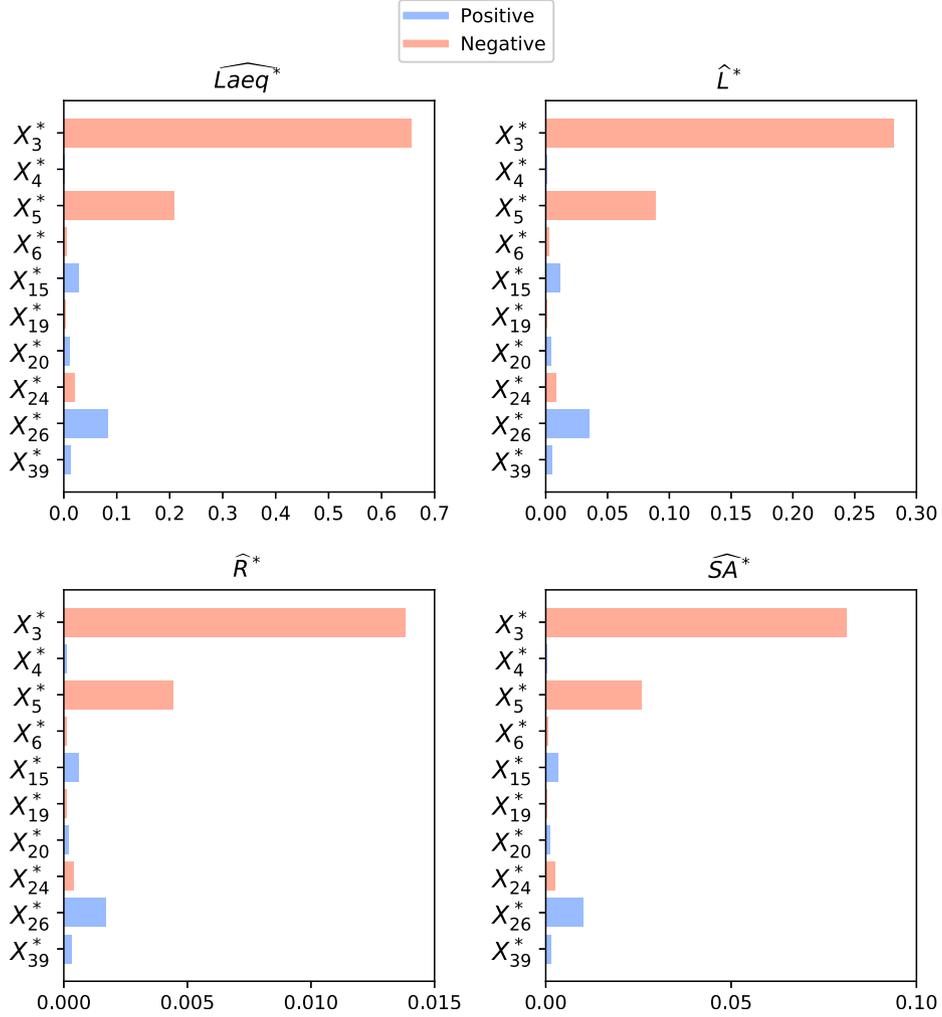

**Fig. 3.** Influence analysis of the input variables considered by the best PUNN model.

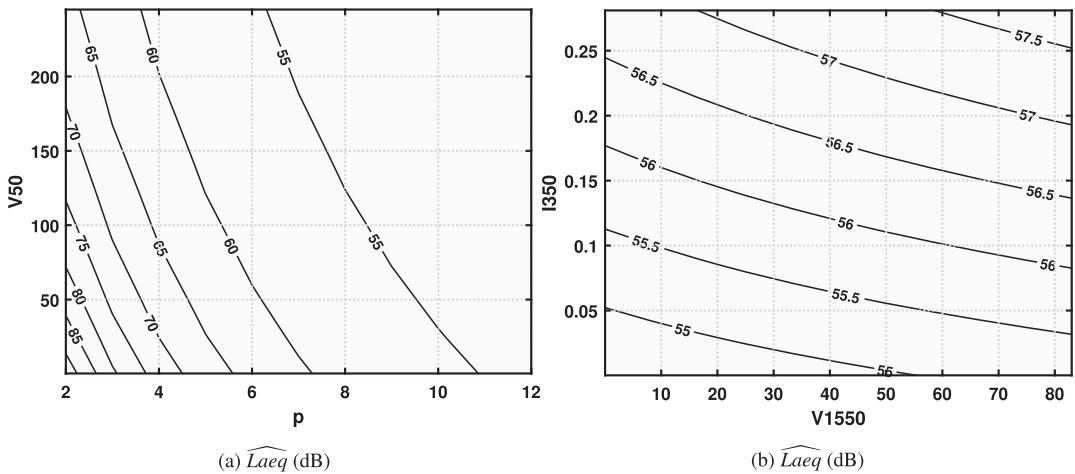

(a) $\widehat{Laeq}$ (dB)　　　　　　　　　　(b) $\widehat{Laeq}$ (dB)

**Fig. 4.** Estimated $\widehat{Laeq}$: (a) in function of $p$ and $V50$, (b) in function of $V1550$ and $I350$, obtained by PUNN model.

Furthermore, it can be appreciated that complex models are able to achieve better results in comparison with simpler models. However, not only is the computational cost higher, but also the number of inputs considered by the model, which, in the case of the best PUNN model is 31, and in the case of the best SUNN model is 37, instead of 10 and 8 which are the inputs considered by the simple models, respectively.

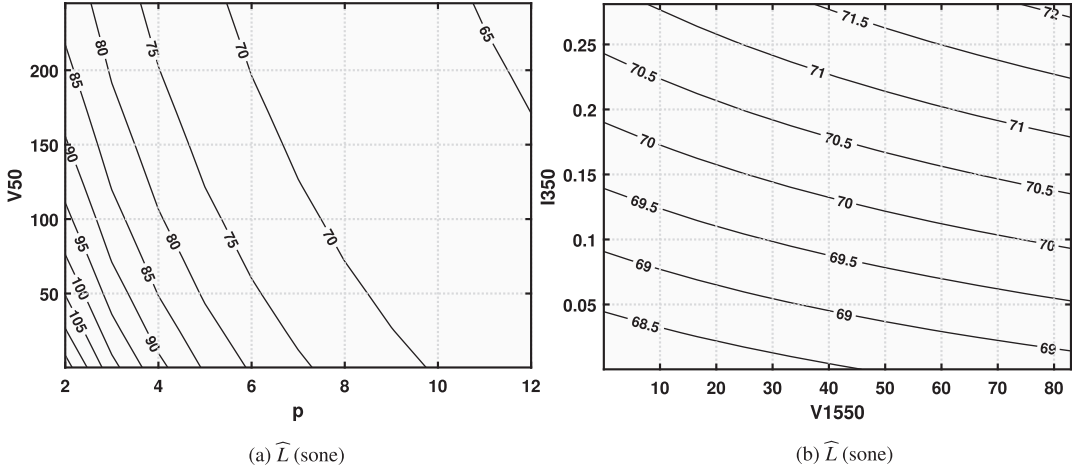

**Fig. 5.** Estimated $\widehat{L}$: (a) in function of $p$ and $V50$, (b) in function of $V1550$ and $I350$, obtained by PUNN model.

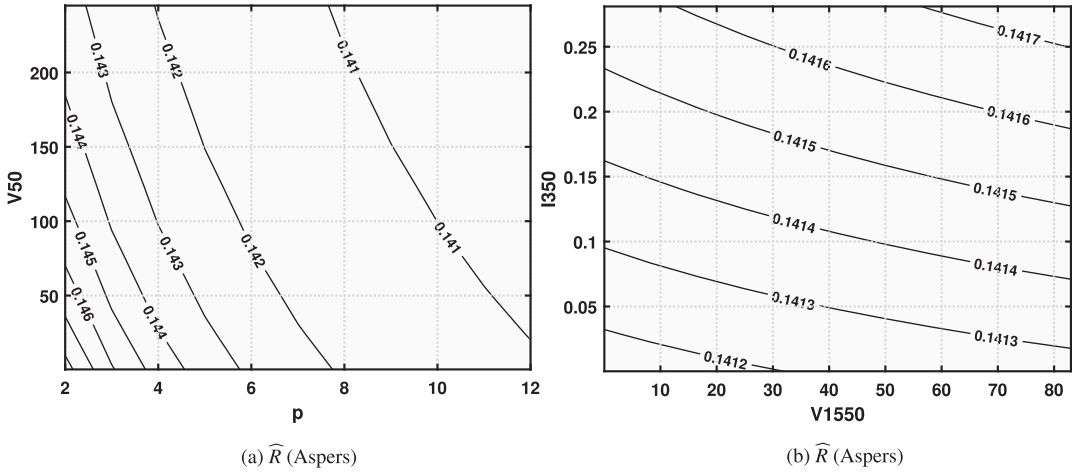

**Fig. 6.** Estimated $\widehat{R}$: (a) in function of $p$ and $V50$, (b) in function of $V1550$ and $I350$, obtained by PUNN model.

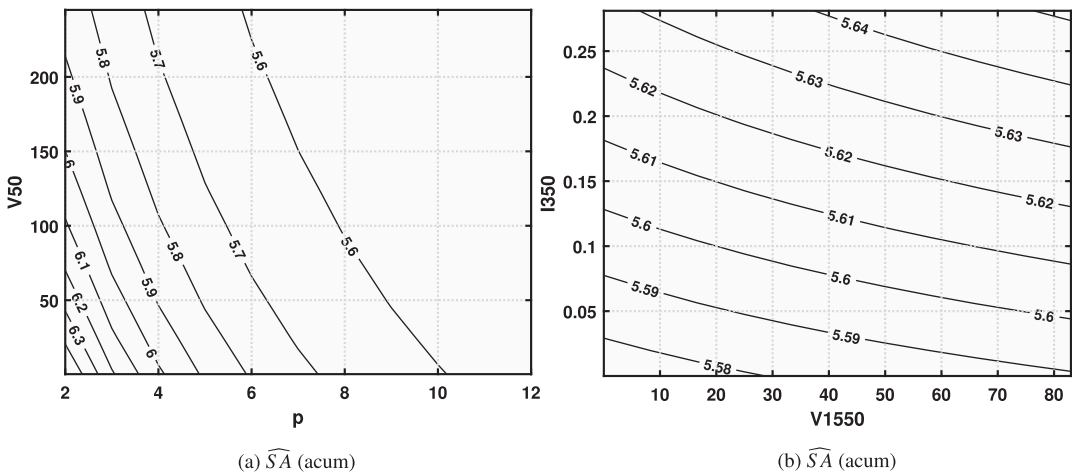

**Fig. 7.** Estimated $\widehat{SA}$: (a) in function of $p$ and $V50$, (b) in function of $V1550$ and $I350$, obtained by PUNN model.

In order to demonstrate the predictive ability of the PUNN and SUNN models, a comparison against the state-of-the-art algorithms has been carried out. For this purpose, the following well-known algorithms Support Vector Regressors (SVR) [30], linear regression (LinearReg) [31], ridge regression (Ridge) [31], Lasso [32] and ElasticNet [33] are used. The results achieved are exposed in terms of MSE and SEP, as well as the number of links of each model, in Table 10. Note that the results for the best PUNN and SUNN models are the same as those shown in Table 9, for visualisation purposes.

**Table 8**
Minimum/difference/maximum value of PUNN model outputs.

| PUNN | $p, V50$ | $V1550, I350$ |
|---|---|---|
| $\widehat{Laeq}$ | 55/35/90 | 55.0/2.5/57.5 |
| $\widehat{L}$ | 65/45/110 | 68.5/3.5/72.0 |
| $\widehat{R}$ | 0.141/0.006/0.147 | $14.12e^{-2}/0.50e^{-3}/14.17e^{-2}$ |
| $\widehat{SA}$ | 5.6/0.8/6.4 | 5.58/0.06/5.64 |

As can be seen, in terms of MSE, the best results are obtained by the SUNN model (29.95), followed by the PUNN model (30.61), whereas, in terms of SEP, linear regression and ridge are the ones achieving the best results (11.42), followed by the SUNN model with a slight difference (11.46). Finally, in terms of complexity, PUNN model is the one with less connections (94 links), followed by the SUNN model (92 links).

## 5. Conclusions

In this study, the use of multitask ANNs to model the acoustic behaviour of an induction motor has been evaluated. For this purpose, the constructive parameter number of poles, and electrical parameters of the feed signal generated by an inverter, were considered.

To obtain a representative set of samples of the noise generated by the induction motor, a large number of experimental tests have been carried out. In this sense, different PWM techniques and diverse combinations of number of poles were used. The following conclusions can be drawn from the results obtained:

1. The best results were obtained by the PUNN model, being the best method to model the noise of the induction motor fed with an inverter, minimizing its approximation error. MSE and SEP values were 38.77 and 13.11, respectively, using just 18 connections. Apart from the outstanding results obtained, this model also considered a reduced number of input parameters, 10.
2. This best model was used to study the noise and acoustic quality of the induction motor, using different combinations of the number of poles and feed signals.
3. The analysis done, stood out that by increasing the fundamental harmonic of the 50 Hz power signal, the sound quality improved, with a low dependence on the number of poles. Moreover, at low speeds ($p = 12$) the fundamental harmonic of 50 Hz does not influence the acoustic quality.
4. ANN models with a more intricate topology and trained by an evolutionary algorithm with a considerably higher number of generations are able to increase the performance of simple models, however, not only is the computational cost higher, but also the number of inputs considered, leading to more complex models sacrificing the interpretability.
5. Multitask ANN models are effective for extracting common characteristics from multiple tasks, and they can be considered outstanding for modelling the noise and acoustic quality of an induction motor powered by an inverter.
6. The whole procedure described in this work has been validated by the competitive results achieved, and can also be applied to any other induction motors, obtaining models with similar characteristics: simplicity and interpretability.

## Acknowledgement

This work has been subsidised by the Spanish Ministry of Economy and Competitiveness (MINECO) and FEDER funds (grant reference TIN2017-85887-C2-1-P) and by the Consejería de Economía,

**Table 9**
Results obtained by the complex models for the four outputs: $Laeq, L, R$ and $SA$.

| | PUNN | | | | | | | | | | |
|---|---|---|---|---|---|---|---|---|---|---|---|
| | MSE | | | | | SEP | | | | | #Links |
| | $\widehat{Global}$ | $\widehat{Laeq}$ | $\widehat{L}$ | $\widehat{R}$ | $\widehat{SA}$ | $\widehat{Global}$ | $\widehat{Laeq}$ | $\widehat{L}$ | $\widehat{R}$ | $\widehat{SA}$ | |
| Mean | 32.53 | 1.12 | 31.34 | $1.74e^{-4}$ | $7.05e^{-2}$ | 11.75 | 0.92 | 3.86 | 4.64 | 2.33 | 54.40 |
| SD | 1.03 | 0.34 | 0.94 | $2.87e^{-6}$ | $4.21e^{-3}$ | 0.19 | 0.12 | 0.06 | 0.04 | 0.07 | 6.14 |
| Best | 30.61 | 1.20 | 29.34 | $1.71e^{-4}$ | $7.07e^{-2}$ | 11.63 | 0.96 | 3.73 | 4.60 | 2.34 | 60.00 |
| | SUNN | | | | | | | | | | |
| | MSE | | | | | SEP | | | | | #Links |
| | $\widehat{Global}$ | $\widehat{Laeq}$ | $\widehat{L}$ | $\widehat{R}$ | $\widehat{SA}$ | $\widehat{Global}$ | $\widehat{Laeq}$ | $\widehat{L}$ | $\widehat{R}$ | $\widehat{SA}$ | |
| Mean | 32.70 | 0.91 | 31.72 | $1.76e^{-4}$ | $6.74e^{-2}$ | 11.65 | 0.83 | 3.88 | 4.66 | 2.28 | 79.30 |
| SD | 1.06 | 0.13 | 0.99 | $7.01e^{-6}$ | $6.45e^{-3}$ | 0.23 | 0.06 | 0.06 | 0.09 | 0.10 | 7.44 |
| Best | 29.95 | 0.79 | 29.09 | $1.76e^{-4}$ | $6.89e^{-2}$ | 11.46 | 0.78 | 3.71 | 4.66 | 2.31 | 92.00 |

**Table 10**
Comparison of the best models achieved with the baseline approaches for the four outputs: $Laeq, L, R$ and $SA$.

| | SVR | | | | | | | | | | |
|---|---|---|---|---|---|---|---|---|---|---|---|
| | MSE | | | | | SEP | | | | | #Links |
| | $\widehat{Global}$ | $\widehat{Laeq}$ | $\widehat{L}$ | $\widehat{R}$ | $\widehat{SA}$ | $\widehat{Global}$ | $\widehat{Laeq}$ | $\widehat{L}$ | $\widehat{R}$ | $\widehat{SA}$ | |
| SVR | 31.81 | 0.94 | 30.81 | $7.58e^{-4}$ | **$5.32e^{-2}$** | 13.87 | 0.85 | 3.78 | 7.22 | **2.02** | 295085 |
| LinearReg | 31.78 | 1.32 | 30.40 | **$1.57e^{-4}$** | $6.46e^{-2}$ | **11.42** | 1.00 | 3.80 | **4.38** | 2.24 | 164 |
| Ridge | 31.69 | 1.31 | 30.31 | $1.59e^{-4}$ | $6.35e^{-2}$ | **11.42** | 1.00 | 3.80 | 4.41 | 2.22 | 164 |
| Lasso | 31.69 | 1.29 | 30.33 | $1.81e^{-4}$ | $6.39e^{-2}$ | 11.72 | 0.99 | 3.80 | 4.70 | 2.23 | 94 |
| ElasticNet | 31.91 | 1.32 | 30.53 | $1.64e^{-4}$ | $6.45e^{-2}$ | 11.54 | 1.00 | 3.81 | 4.49 | 2.24 | 137 |
| PUNN | 30.61 | 1.20 | 29.34 | $1.71e^{-4}$ | $7.07e^{-2}$ | 11.63 | 0.96 | 3.73 | 4.60 | 2.34 | **60** |
| SUNN | **29.95** | **0.79** | **29.09** | $1.76e^{-4}$ | $6.89e^{-2}$ | 11.46 | **0.78** | **3.71** | 4.66 | 2.31 | 92 |

Conocimiento, Empresas y Universidad de la Junta de Andalucía (grant reference UCO-1261651). David Guijo-Rubio's research has been subsidised by the FPU Predoctoral Program (Spanish Ministry of Education and Science), grant reference FPU16/02128.

## Appendix A. Supplementary data

Supplementary data associated with this article can be found, in the online version, at https://doi.org/10.1016/j.apacoust.2020.107332.